\documentclass{article}

\usepackage{amsmath,amsfonts,bm}

\def\eqref#1{equation~\ref{#1}}

\def\1{\bm{1}}

\def\rmA{{\mathbf{A}}}

\def\rmC{{\mathbf{C}}}

\def\rmE{{\mathbf{E}}}

\def\rmL{{\mathbf{L}}}

\def\rmP{{\mathbf{P}}}
\def\rmQ{{\mathbf{Q}}}

\def\rmX{{\mathbf{X}}}
\def\rmY{{\mathbf{Y}}}

\def\vmu{{\bm{\mu}}}

\def\vp{{\bm{p}}}

\def\vv{{\bm{v}}}

\def\vz{{\bm{z}}}

\DeclareMathAlphabet{\mathsfit}{\encodingdefault}{\sfdefault}{m}{sl}
\SetMathAlphabet{\mathsfit}{bold}{\encodingdefault}{\sfdefault}{bx}{n}

\def\gM{{\mathcal{M}}}

\def\gP{{\mathcal{P}}}

\def\sR{{\mathbb{R}}}

\newcommand{\R}{\mathbb{R}}

\newcommand{\softmax}{\mathrm{softmax}}

\DeclareMathOperator*{\argmax}{arg\,max}

\usepackage{colortbl}
\usepackage[preprint]{neurips_2024}
\usepackage{optidef}
\usepackage{algpseudocode, algorithm}
\usepackage[colorlinks=True]{hyperref}
\usepackage{url}
\usepackage[capitalize]{cleveref}
\crefname{equation}{}{}
\usepackage{mathtools}
\usepackage{amsmath,amsfonts,amssymb,amsthm,commath}
\usepackage{subcaption}

\DeclarePairedDelimiter\autobracket{(}{)}
\newcommand{\p}[1]{\autobracket*{#1}}
\def\vnu{{\bm{\nu}}}
\def\1{\bm{1}}
\newcommand{\ip}[2]{\langle#1, #2\rangle}

\usepackage[inline, shortlabels]{enumitem}

\usepackage[utf8]{inputenc} %
\usepackage[T1]{fontenc}    %
\usepackage{hyperref}
\hypersetup{
colorlinks = true,
linkcolor = blue, 
filecolor = magenta,
urlcolor = blue,
citecolor = blue 
}
\usepackage{url}            %
\usepackage{booktabs}       %
\usepackage{amsfonts}       %
\usepackage{nicefrac}       %
\usepackage{microtype}      %
\usepackage{xcolor}         %

\definecolor{lightblue}{RGB}{173,216,230}
\definecolor{lightgreen}{RGB}{144,238,144}
\definecolor{lightred}{RGB}{255,182,193}

\definecolor{first}{RGB}{102,194,165}
\definecolor{second}{RGB}{140,209,182}
\definecolor{third}{RGB}{178,224,199}
\definecolor{fourth}{RGB}{216,239,216}
\definecolor{fifth}{RGB}{240,247,237}

\definecolor{orangeFirst}{RGB}{252, 141, 98}   %
\definecolor{orangeSecond}{RGB}{255, 180, 150} %
\definecolor{orangeThird}{RGB}{255, 200, 175}  %
\definecolor{orangeFourth}{RGB}{255, 220, 200} %
\definecolor{orangeFifth}{RGB}{255, 240, 225}  %

\title{Adapting Language Models via Token Translation}

\author{%
  Zhili Feng \\
  Carnegie Mellon University \\
  \And
  Tanya Marwah\\
  Carnegie Mellon University \\
  \And
  Nicol\`{o} Fusi\\
  Microsoft Research \\ 
  \And
  David Alvarez-Melis\\
  Microsoft Research \\
  \And
  Lester Mackey\\
  Microsoft Research
}

\begin{document}

\maketitle

\begin{abstract}
Modern large language models use a fixed tokenizer to effectively compress text drawn from a source domain. However, applying the same tokenizer to a new target domain often leads to inferior compression, more costly inference, and reduced semantic alignment. To address this deficiency, we introduce Sparse Sinkhorn Token Translation (S2T2). S2T2 trains a tailored tokenizer for the target domain and learns to translate between target and source tokens, enabling more effective reuse of the pre-trained next-source-token predictor. In our experiments with finetuned English language models, S2T2 improves both the perplexity and the compression of out-of-domain protein sequences, outperforming direct finetuning with either the source or target tokenizer. In addition, we find that token translations learned for smaller, less expensive models can be directly transferred to larger, more powerful models to reap the benefits of S2T2 at lower cost.
\end{abstract}

\section{Introduction}
Modern large language models (LLMs) are typically trained in two stages. 
First a tokenizer is trained to map commonly occurring character sequences in the training data into vocabulary units known as \emph{tokens}.   Next, all training text is tokenized, i.e., translated into this token vocabulary, and a model is trained to predict the next token given a context of preceding tokens. 
The tokenizer can be viewed as an initial compressor of input bytes \citep{gage1994new} that 
significantly shortens text drawn from the training domain 
and arguably improves the training dynamics \citep{rajaraman2024toward}. 
Despite its widespread adoption, this two-stage procedure suffers from a key failing: When faced with text from a new target domain, compression quality drops, context length and inference costs  increase, and learned semantic alignment deteriorates. 
This effect is especially evident when modern LLMs (trained predominantly on English and code) are used to reason about molecular sequences like proteins. Such sequences are commonly represented using the Latin-script alphabet, but the meaning and frequency of each substring differ significantly their natural language counterparts, resulting in semantic misalignment. 

To tackle the analogous alignment problem for low-resource languages, \citet{remy2024transtokenization} proposed to use 
\texttt{fast\_align} \citep{dyer2013simple}, 
an expectation-maximization algorithm that requires parallel data from the training and target domains. 

This approach shows promising results, but
for many target domains, 
parallel training data is difficult or impossible to gather. %
For example, there is no agreed-upon parallel translation between protein sequences and natural language. 

In this work, we propose a \textit{\textbf{S}parse \textbf{S}inkhorn \textbf{T}oken \textbf{T}ranslation (S2T2)} algorithm that does not require parallel data. Instead, S2T2 learns a translation between training domain tokens and new target domain tokens just using a sample data from the target domain and the pretrained LLM weights. 
After training a tokenizer on the target domain, S2T2  
translates each target-domain token into a (sparse) distribution over training-domain tokens, uses the  pretrained LLM to predict the next training-domain token, and  translates that training-domain token back into a (sparse) distribution over target-domain tokens. 
In our experiments with English LLMs, we find that 
\begin{enumerate}[leftmargin=*]
	\item S2T2 provides an effective initialization for continual finetuning on protein sequences, yielding both better compression and better perplexity than direct finetuning of the pretrained model, and %
	\item S2T2 enables \textit{weak-to-strong model transferability}: Translations learned for smaller, less expensive models can be transferred to larger, more powerful models to reap the benefits at lower cost. %
\end{enumerate}

\section{Translating Tokens with Sparse Sinkhorn}

\begin{figure*}[t!]
    \centering
    \begin{subfigure}[t]{0.45\textwidth}
    \centering
        \includegraphics[height=5cm]{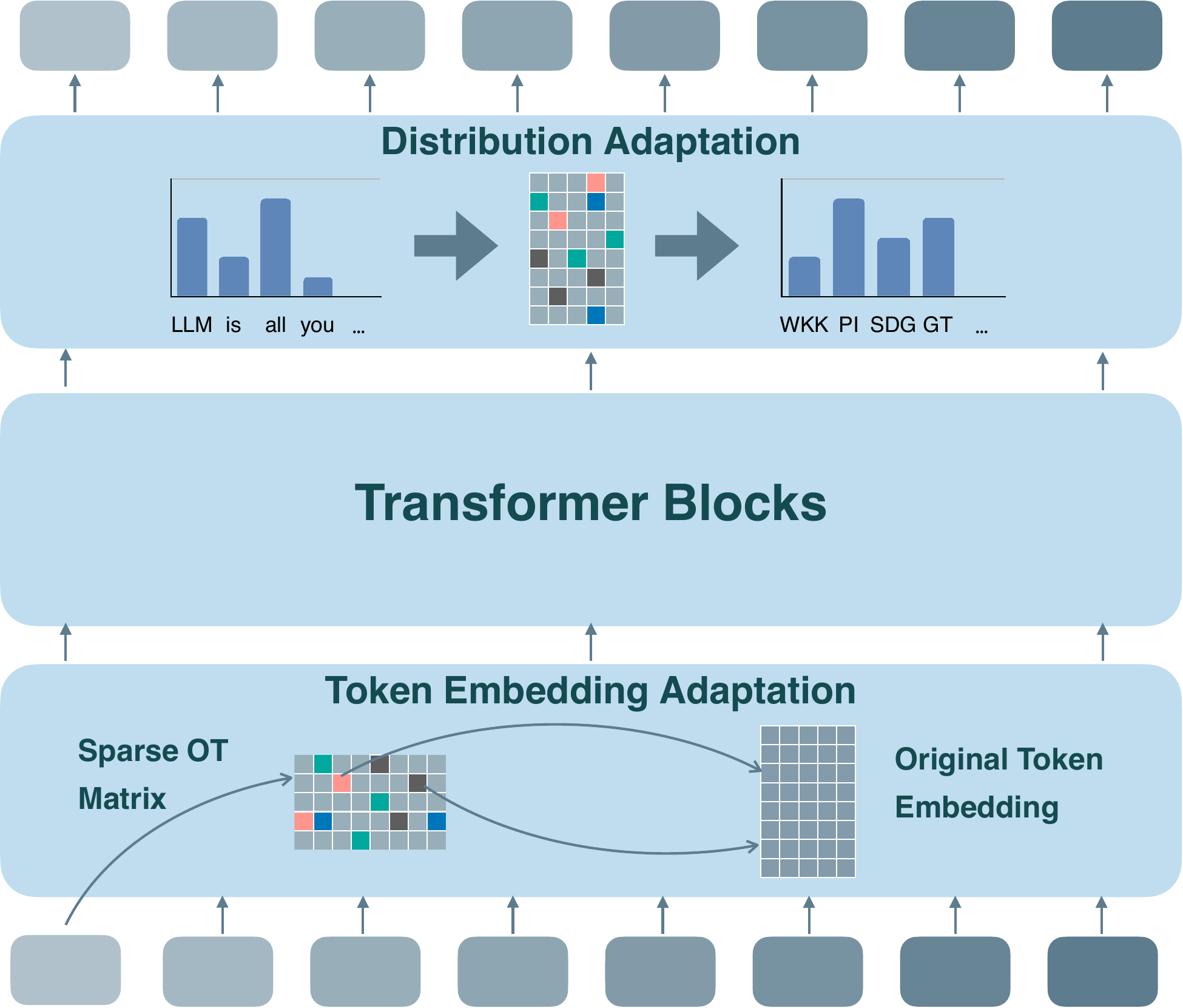}
    \end{subfigure}%
    ~
    \raisebox{0.5cm}{
    \begin{subfigure}[t]{0.55\textwidth}
    \centering
        \includegraphics[height=4cm]{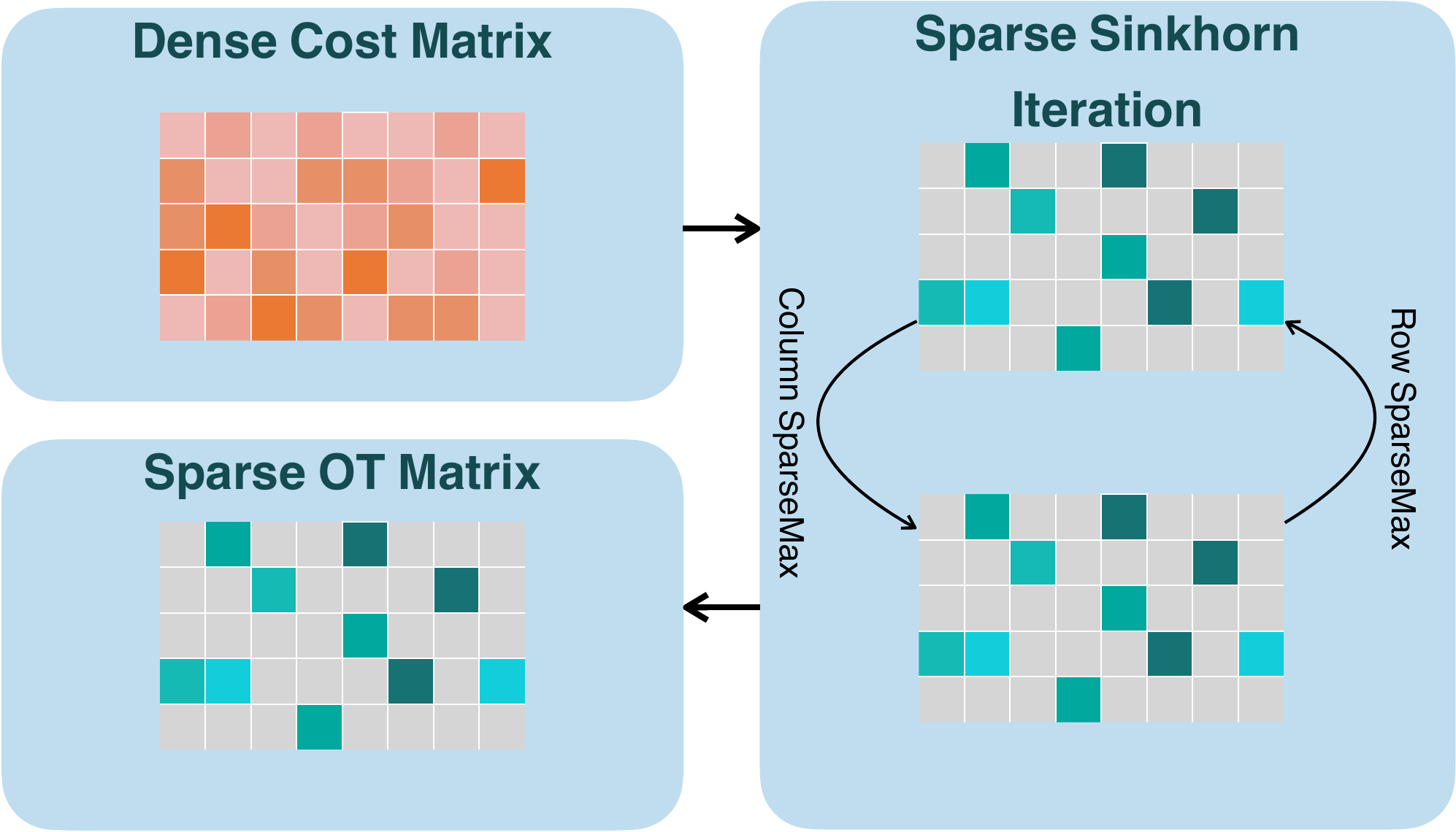}
    \end{subfigure}
    }
    \caption{Overview of S2T2. \textbf{Left}: S2T2 injects a weight-tied sparse optimal transport (OT) layer in both the token embedding and language model head. The input tokens will be encoded based on a sparse convex combination of the original token embeddings and decoded by a sparse combination of the original language model head.
    \textbf{Right}: The sparse OT matrix is obtained by iteratively projecting a dense cost matrix along its rows and columns. The dense cost matrix is updated by backpropogation.}
    \label{fig:overview}
\end{figure*}

Consider a pretrained LLM $\mathcal{M}$ with vocabulary size $v$, embedding matrix $\rmE\in\mathbb{R}^{v\times d}$, and language model head $\rmL\in\mathbb{R}^{v\times d}$.  
For a given input sequence encoded as a matrix $\rmX\in\{0, 1\}^{s\times v}$ in which each row is a one-hot vector representing a training-domain token, 
$\rmX\rmE\in\sR^{s\times d}$ represents the sequence of (soft) embeddings, and 
the predicted next token is given by 
\begin{align}\label{eq:next_token}
\gM(\rmX) = \argmax_{i\in[v]} \softmax(\rmL h(\rmX\rmE))_i \in \{0,1\}^v %
\end{align}
where $h: \sR^{s\times d}\to\sR^d$ maps an embedding sequence into a single vector, the internal representation of the next token.

\newcommand{\simplex}[1]{\Delta_{#1-1}}
Consider also a dataset $D$ %
drawn from a new target domain,  and let $u$ be the vocabulary size of a new tokenizer trained on $D$. 
For given marginal distributions over training and target tokens $\vmu\in\simplex{v}$ and $\vnu\in\simplex{u}$, we define the constraint set $C(\vmu,\vnu)=\{\rmP\in [0,1]^{v\times u}:\rmP\bm{1}=\vmu, \rmP^\top\1=\vnu \}$.   
S2T2 finds a joint probability matrix $\rmP\in C(\vmu,\vnu)$ and defines a new target-domain LLM $\gM'$ with 
embedding matrix $\rmE'=\p{\rmP^\top\odot (1/\vmu)}\rmE\in\sR^{u\times d}$ and language head $\rmL'=\p{\rmP\odot (1/\vnu)}^\top\rmL\in\sR^{u\times d}$ substituted for $(\rmE, \rmL)$ in \cref{eq:next_token}.
Here, $\rmA\odot\vv$ represents a Hadamard product broadcasted along the last dimension. It is crucial to perform such a Hadamard product, since we want the new token embedding and old token embedding to be on the same scale. More generally, one could use different $\rmP$ matrices to translate $\rmE$ and $\rmL$, but we focus on a single $\rmP$ here for simplicity. %
An overview of S2T2 can be in \cref{fig:overview}.

\subsection{Finding $\rmP$ via Sparse Sinkhorn}
Since it is difficult to directly parameterize a joint probability matrix $\rmP\in C(\vmu,\vnu)$, we 
instead maintain a dense weight matrix $\rmC\in\sR^{v\times u}$  and recover $\rmP$
as the solution to the following two equivalent optimization problems.

\begin{minipage}{0.45\linewidth}
	\begin{mini}|s|
		{\rmP'}{\frac{1}{2}\|\rmP'-\rmC\|_F^2}{}{}
		\label{eq:opt1}
            \addConstraint{\rmP'\in C(\vmu, \vnu)}{}
	\end{mini}
\end{minipage}
\begin{minipage}{0.45\linewidth}
	\begin{mini}|s|
		{\rmP'}{\ip{-\rmC}{\rmP'}+\frac{1}{2}\|\rmP'\|_F^2}{}{}
		\label{eq:opt2}
            \addConstraint{\rmP'\in C(\vmu, \vnu)}{}
	\end{mini}
\end{minipage}

Notice that \cref{eq:opt2} is the $\ell_2$ constrained optimal transport problem, which is known to generate sparse solutions \citep{essid2018quadratically,peyre2019computational}. 
Moreover, since $C=C_1\cap C_2$ for the convex sets $C_1=\{\rmP\in\sR_+^{v\times u}, \rmP\1=\vmu\}$ and $C_2=\{\rmP\in\sR_+^{v\times u}, \rmP^\top\1=\vnu\}$, these problems can be solved using iterative Dykstra's projections \citep{boyle1986method}, a Sinkhorn-like algorithm via  with guaranteed convergence (see \cref{alg:sparse_sinkhorn}). 

In every Sinkhorn iteration, we solve a set of $\ell_2$ projections onto a probability simplex. This optimization problem enjoys an efficient backpropogation computation \citep{martins2016softmax}. A small caveat is that we are not always projecting onto the unit simplex but rather onto a scaled simplex, so the optimization is modified accordingly in  \cref{alg:sparsemax}.

\begin{algorithm}
\caption{\textsc{Sparse Sinkhorn Iteration}}
\label{alg:sparse_sinkhorn}
\begin{algorithmic}[1]
\Require Weight matrix $\rmC\in\R^{v\times u}$
\State $\rmP\gets \bm{0}^{v\times u}$, $\rmQ\gets \bm{0}^{v\times u}$, $\rmX_0\gets\rmC$
\For{$k=0,\ldots, n$}
	\State $\rmY_k \gets \gP_{C_1}(\rmX_k+\rmP_k)$, where $\gP_{C_1}$ applies \textsc{Sparsemax} with scale $\vmu_i$ to each row $i$. %
	\State $\rmP_{k+1} \gets \rmX_k+\rmP_k-\rmY_k$
	\State $\rmX_{k+1} \gets \gP_{C_2}(\rmY_k+\rmQ_k)$, where $\gP_{C_2}$ applies \textsc{Sparsemax} with scale $\vnu_j$ to each column $j$. %
	\State $\rmQ_{k+1}\gets \rmY_k+\rmQ_k-\rmX_{k+1}$
\EndFor
\State \Return $\rmX_{n+1}$
\end{algorithmic}
\end{algorithm}

\begin{algorithm}
\caption{\textsc{Sparsemax}}
\label{alg:sparsemax}
\begin{algorithmic}[1]
\Require $\vz\in\R^K$, scale $\alpha$
\State Sort $\vz$ as $\vz_{(1)}\geq\cdots\geq\vz_{(K)}$
\State Find $k(\vz)=\max\left\{k\in[K]: \alpha+k\vz_{(k)}>\sum_{j\leq k}\vz_{(j)}\right\}$
\State Let $\tau(\vz)=\frac{\sum_{j\leq k}\vz_{(j)}-\alpha}{k(\vz)}$
\State \Return $\vp$ where $\vp_i=\max\{\vz_i-\tau(\vz), 0\}$
\end{algorithmic}
\end{algorithm}

To learn our token translation, we initialize the weight matrix $\rmC$ by setting each entry to be $1/v$, obtain the joint probability matrix $\rmP$ by applying \cref{alg:sparse_sinkhorn} to $\rmC$, and perform a normal forward pass using $\rmP$. During the backward pass, we differentiate through the Sinkhorn iteration and update $\rmC$ directly. In practice, we find that iterating $3$ times is enough to generate an effective sparse $\rmP$.

\section{Experiment}
We conduct experiments on the UniRef50 \citep{suzek2015uniref} protein sequence dataset using the OLMo-1B English LLM \citep{Groeneveld2023OLMo} with batch size $16$ and context length of $512$. 
The training domain tokens in our experiment are bytes (single characters), and the target domain tokenizer is a new Byte-Pair Encoding (BPE) tokenizer \citep{gage1994new} trained on UniRef50 with vocabulary size $512$. 
The new tokenizer reduces the length our protein sequences by  a factor of $1.82\times$ on average. %
This will in turn have sizable impact on the standard measure of model compression, bits-per-byte (BpB)  \citep[see][for  details on calculating BpB]{biderman2024lessons}. To control the sparsity level of $\rmP$, we add an entropy regularizer $\alpha H(\rmP)$ to the next token prediction loss with larger $\alpha$ encouraging smaller entropy and hence sparser $\rmP$. 
Unless otherwise specified, $\alpha=0$. 

We compare with four baseline methods:
\begin{enumerate*}[series = tobecont, itemjoin = \quad]
	\item Training an  \emph{unconstrained} translator $\rmP$ followed by whole-model finetuning.
	\item Training a dense probabilistic translator $\rmP$ (using \textsc{SoftMax} in place of \textsc{SparseMax}) followed by whole-model finetuning.
	\item Finetuning the model directly using the original OLMo tokenizer. 
	\item Finetuning the model with the new tokenizer, resizing the embedding matrix $\rmE$ and language model head $\rmL$ by truncation.
\end{enumerate*}

\paragraph{Training details.} We always train with AdamW \citep{loshchilov2018decoupled}. When training $\rmP$, we use a learning rate of $10^{-3}$ (except for our model transfer experiments, which use $2\times 10^{-5}$) and no weight decay; when finetuning the whole model, we always use learning rate of $2\times 10^{-5}$ with $0.01$ weight decay. We follow the convention of training with BFloat16, $\beta_1=0.9, \beta_2=0.95,$ and $\varepsilon=10^{-5}$. We always use the cosine annealing scheduler with $20\%$ linear warm-up steps and decay to $10\%$ of the learning rate.
We train $\rmP$ and finetune the whole model with $2000$ steps. 

\begin{figure*}
	\centering
	\includegraphics[scale=0.4]{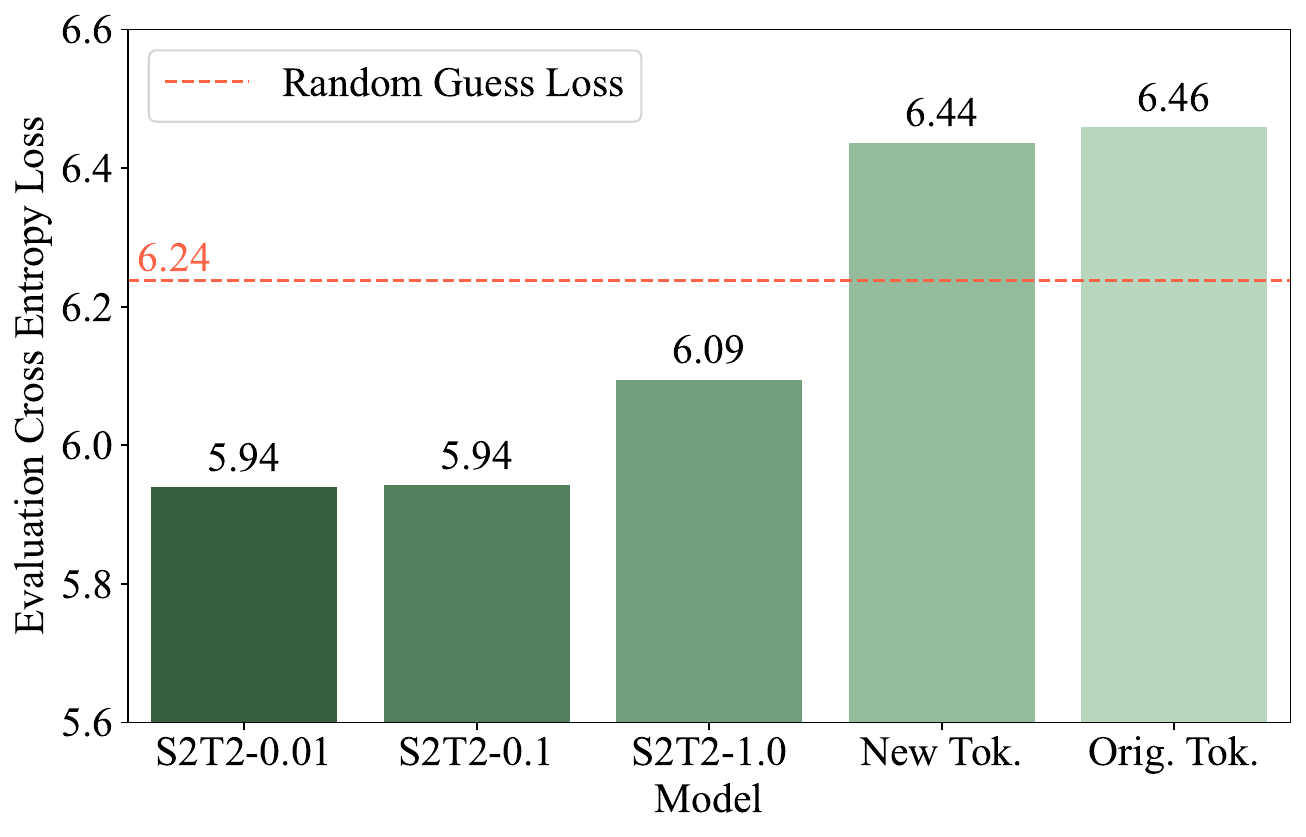}
	\caption{Evaluation loss after initializing OLMo-7B with token translator $\rmP$ learned from OLMo-1B.  Along the x-axis, $\text{S2T2-}\alpha$ represent S2T2 with the $\alpha$-entropy regularizer that controls the sparsity of $\rmP$. \textbf{New Tok.} is OLMo-7B with the new tokenizer and truncated $\rmE, \rmL$; \textbf{Orig Tok.} is OLMo-7B with the original tokenizer. The red dashed line is the loss when you \textbf{randomly guess} the next token.}
 \label{fig:model_transfer}
\end{figure*}

Remarkably, \cref{tab:performance} shows that simply initializing with S2T2 produces better language model quality (as measured by perplexity) and compression (as measured by BpB) than whole-model finetuning with the original tokenizer (baseline 3). 
Note that baseline 3 has much worse BpB due to its longer sequence length, further motivating the usage of a tailored tokenizer.  
In addition, S2T2 initialization outperforms both dense Sinkhorn and unconstrained token translation in both metrics. 
Moreover, after finetuning, S2T2 also improves upon the perplexity and BpB of baseline 4, direct finetuning with a new tokenizer.  
\cref{fig:model_transfer} shows that the translator $\rmP$ learned using OLMo-1B can also be directly transferred to the more expensive model, OLMo-7B, yielding significantly better performance than random guessing or OLMo-7B with its original tokenizer or the new tokenizer with truncated embedding matrix and language model head.

\begin{table}[htbp]
\centering
\caption{Performance on UniRef50 evaluation set, measured by perplexity (\textbf{perp.}) and bits-per-byte (\textbf{BpB}). \textbf{Plain $\rmP$}: Unconstrained $\rmP$. \textbf{CFT}: Continual finetuning, initialized from the learned $\rmP$. \textbf{FT orig. tok.}: Finetuning with the original tokenizer. \textbf{FT new tok.}: Finetuning with the new tokenizer. }
\resizebox{\textwidth}{!}{
\begin{tabular}{lcc|cc|cc|c|c}
\toprule
  &  \textbf{Plain $\rmP$} & \textbf{+ CFT} & \textbf{Sinkhorn $\rmP$} & \textbf{+ CFT} & \textbf{S2T2} & \textbf{+ CFT} & \textbf{FT orig. tok.} & \textbf{FT new tok.} \\
  \midrule
  \textbf{Perp.} & 174.20 & \cellcolor{second} 130.44 & 167.74 & \cellcolor{third}136.12 & \cellcolor{third} 144.03 & \cellcolor{first}118.78 & 151.05 & \cellcolor{second}130.56 \\
  \midrule
  \textbf{BpB} & 4.09 & \cellcolor{orangeSecond}3.86 & 4.06 & \cellcolor{orangeSecond}3.89 & \cellcolor{orangeThird}3.94 & \cellcolor{orangeFirst}3.78 & 7.24 & \cellcolor{orangeSecond}3.86 \\
\bottomrule
\end{tabular}
}
\label{tab:performance}
\end{table}

\section{Conclusion}
We proposed S2T2 as a token translation technique for continual finetuning of LLMs on out-of-distribution data and demonstrate its effectiveness on protein sequence modeling. As a next step, we plan to expand this framework to adapt to other modalities such as code and images. 
Another natural extension is to combine the training and target token vocabularies to produce an effective ``multidomain'' LLM. %

\section{Acknowledgement}
This work was done during Zhili Feng's and Tanya Marwah's internship at Microsoft Research New England.

\bibliography{citations}
\bibliographystyle{plainnat}

\end{document}